\documentclass[twocolumn,10pt]{article}

\usepackage{arxiv}

\usepackage[utf8]{inputenc} % allow utf-8 input
\usepackage[T1]{fontenc}    % use 8-bit T1 fonts
\usepackage{url}            % simple URL typesetting
\usepackage{graphicx}
\usepackage{booktabs}       % professional-quality tables
\usepackage{amsfonts}       % blackboard math symbols
\usepackage{amssymb}        % provides \varnothing used in algorithms
\usepackage{microtype}      % microtypography

\usepackage{xspace} % loaded by cvpr.sty in the CVPR version
\usepackage{mathtools}
\usepackage{tabularx}
\usepackage{multirow}
\usepackage{placeins}
\usepackage{algorithm}
\usepackage{algpseudocode}
\usepackage{tikz}
\usetikzlibrary{arrows.meta,positioning,fit}
\definecolor{draftblue}{RGB}{36,91,135}
\newcommand{\method}{\textnormal{\textsc{PlaySmith}}\xspace}

\newcommand{\yes}{\ensuremath{\checkmark}}
\newcommand{\no}{\ensuremath{\times}}

\newcommand{\E}{\mathbb{E}}

\DeclareMathOperator*{\argmin}{arg\,min}
\DeclareMathOperator*{\argmax}{arg\,max}

\usepackage{listings}
\usepackage{natbib}
\setcitestyle{numbers,sort&compress}
\usepackage[colorlinks=true, allcolors=black]{hyperref}

\title{From Watching to Playing: Auto-Playing Video Games\\with a Self-Constructed Code World Model}

\author{
    \textbf{Yao Tong}$^{1}$\thanks{Equal contribution.}\\
    Hangzhou Institute for Advanced Study\\
    University of Chinese Academy of Sciences\\
    Hangzhou, China\\
    \texttt{tongyao261@mails.ucas.ac.cn} \\
    \And
    \textbf{Jiaxin Qi}$^{2}$\footnotemark[1]\\
    Computer Network Information Center\\
    Chinese Academy of Sciences\\
    Beijing, China\\
    \texttt{jxqi@cnic.cn} \\
    \authorrowbreak
    \textbf{Yan Cui}$^{1}$\\
    Hangzhou Institute for Advanced Study\\
    University of Chinese Academy of Sciences\\
    Hangzhou, China\\
    \texttt{cuiyan.ch@gmail.com} \\
    \And
    \textbf{Yuhua Zheng}$^{1}$\\
    Hangzhou Institute for Advanced Study\\
    University of Chinese Academy of Sciences\\
    Hangzhou, China\\
    \texttt{zhengyuhua@ucas.ac.cn} \\
    \authorrowbreak
    \textbf{Jianqiang Huang}$^{2}$\thanks{Corresponding author.}\\
    Computer Network Information Center\\
    Chinese Academy of Sciences\\
    Beijing, China\\
    \texttt{jqhuang@cnic.cn} \\
}

\renewcommand{\headeright}{Preprint}
\renewcommand{\undertitle}{Preprint}
\renewcommand{\shorttitle}{From Watching to Playing}

\hypersetup{
pdftitle={From Watching to Playing: Auto-Playing Video Games with a Self-Constructed Code World Model},
pdfsubject={cs.AI, cs.CV},
pdfauthor={Yao Tong, Jiaxin Qi, Yan Cui, Yuhua Zheng, Jianqiang Huang},
pdfkeywords={Code World Model, Vision-Language Model, Game Agent, Program Synthesis},
}

\begin{document}

\maketitle
% Title-block notes (equal contribution / corresponding author) use their own
% footnote symbols; restart body footnote numbering at 1.
\setcounter{footnote}{0}

% !TEX root = ../main.tex
\begin{abstract}

Vision-language model (VLM) agents can play video games by selecting
actions from successive screenshots, but calling a VLM at each decision
step incurs prohibitive time and token costs. Compiling a VLM's game
knowledge into reusable code would amortize these costs, yet raw visual
gameplay entangles autonomous physics with player actions, obscuring
the transition dynamics a compiled model needs. To address both challenges,
we present \method, an agent harness that constructs three modules from
visual gameplay. First, a visual Observer detects game objects and tracks
them across frames with persistent identity. Second, action-conditioned
displacement (ACD) links actions to object motion, guiding construction
of a stateful Predictor that distinguishes action-driven motion from
autonomous physics.
Third, a controller uses the Predictor to evaluate candidate action
sequences multiple steps ahead and executes the selected action.
Across eight 2D browser games, \method{} achieves competitive scores
relative to direct VLM gameplay using the same backbone, while consuming
no VLM tokens during evaluation episodes per game and reducing mean
per-step latency from 4.526 to 0.293 seconds. These results demonstrate
that a self-constructed code world model can be more practical than direct
VLM play in settings requiring repeated deployment.
\end{abstract}

\begin{figure}[t]
\centering
\includegraphics[width=0.98\columnwidth]{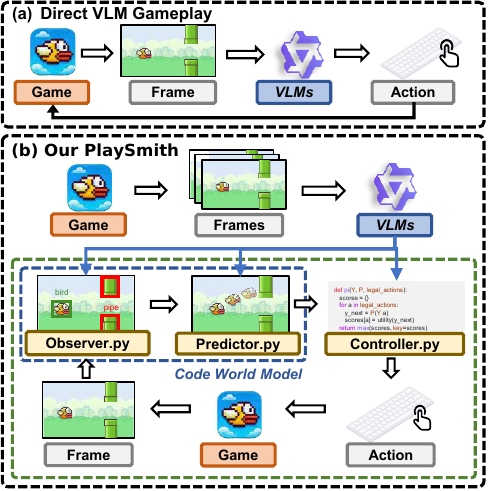}
\caption{\textbf{VLM use in game agents.}
(a) Existing VLM-based agents call the model at each frame. (b) PlaySmith calls
it during development and runs the resulting code during gameplay.}
\label{fig:teaser}
\end{figure}

% !TEX root = ../main.tex
\section{Introduction}
\label{sec:intro}
Modern vision-language models (VLMs) have strong visual understanding
and multimodal reasoning capabilities. They can recognize objects,
interpret spatial relationships, and use visual evidence to decide how
to act. These abilities let VLM agents play video games by reading
successive screenshots and selecting keyboard or pointer actions, as
demonstrated by prior visual agents~\cite{tan2024cradle,han2026vista}.
They can do so without direct access to the game's internal state or
rules, using feedback from successive screenshots to infer what changed
and adjust their subsequent choices.

Despite their strong game-playing capabilities, calling a VLM at every
decision (Fig.~\ref{fig:teaser}) consumes time and tokens even when the
same mechanic governs successive decisions. A code world model (CWM)
offers a way to amortize this reasoning: once it captures a game's
mechanics, it can be reused across play sessions at low marginal cost.
Studies have shown that code world models can capture physical regularities~\cite{wang2026codeasworld}
or simulate action-dependent transitions for planning~\cite{tang2024worldcoder,dainese2024gifmcts,piriyakulkij2025poeworld}, which demonstrates that using a CWM to carry a VLM's knowledge of the game world is feasible.
However, building one from visual gameplay is difficult because action effects
and autonomous dynamics appear together. Prior code world models typically
rely on privileged game states, simulator APIs, or clean vector traces,
leaving direct induction from raw visual pixels largely unexplored.

To bridge this gap, we introduce \method, a game harness that constructs
three modules from visual gameplay (Fig.~\ref{fig:teaser}). First, a
visual Observer detects game objects and associates them across frames,
producing histories with persistent identity. Second, the harness pairs
executed actions with object displacements in action-conditioned
displacement (ACD) records. These records distinguish input-driven
changes from autonomous dynamics and guide construction of a stateful
Predictor. Together, the Observer and Predictor form the CWM. Third, a
controller evaluates candidate action sequences through the Predictor,
executes the chosen action, and updates its state from the next
observation, without further language-model calls.

Compared with direct VLM play using the same backbone, \method{} achieves
stronger gameplay performance while requiring no VLM calls during evaluation
and substantially reducing per-step latency. Under a common frozen Observer
and controller and a comparable Predictor-construction budget, it also
outperforms construction recipes adapted from prior work in both immediate
and multi-step prediction. These results show that a self-constructed CWM
can combine accurate dynamics with efficient repeated deployment.

We make the following three contributions:
\begin{itemize}
\item We introduce a visual-to-code harness that constructs an executable
CWM from tracked gameplay and pairs it with a controller that plays
without online VLM calls.
\item We evaluate object tracking, multi-step prediction, and control
across eight games, including a matched comparison with direct VLM
gameplay and measurements of construction cost.
\item We show through ablations that aligning executed actions with
object displacements improves multi-step prediction and downstream
control, clarifying the role of ACD in constructing the Predictor.
\end{itemize}

\begin{figure*}[t]
\centering
\includegraphics[width=\textwidth]{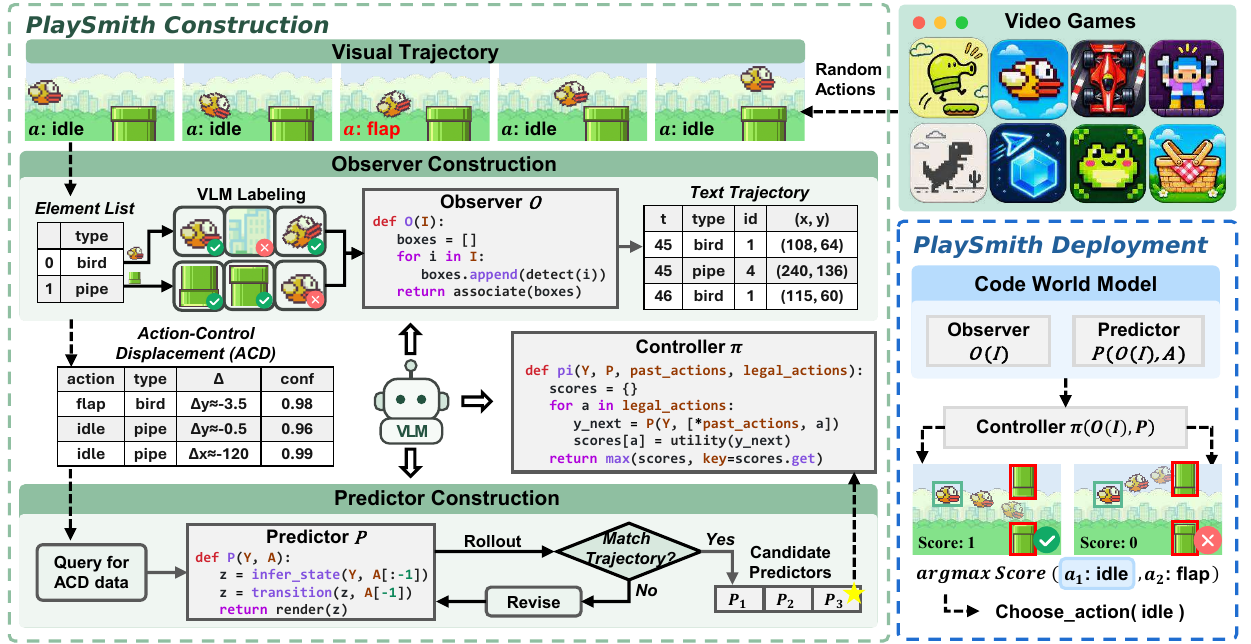}
\caption{\textbf{CWM construction and deployment.}
A VLM labels gameplay frames to build the Observer, and action-conditioned
displacements guide Predictor synthesis and revision. At deployment, the
controller scores candidate actions using predicted outcomes.}
\label{fig:method}
\end{figure*}

% !TEX root = ../main.tex
\section{Related Work}
\label{sec:related}

\paragraph{Game-playing agents.}
MineDojo uses Minecraft to evaluate open-ended embodied tasks and
provides multimodal resources for agent development~\cite{fan2022minedojo}.
Voyager, Cradle, VISTA, SIMA 2 and Odysseus focus on skill acquisition or action
selection~\cite{wang2023voyager,tan2024cradle,han2026vista,simateam2025sima2,shi2026odysseus}.
Code to Control and Compiled Agency develop executable
controllers~\cite{ahmed2026codetocontrol,huang2026compiledagency}.
Visual agents can invoke a foundation model during play, while executable
controllers move action selection into reusable code. \method{} also
constructs a stateful Predictor, so its learned dynamics can be reused across
episodes without online VLM calls. Evaluating the Predictor on held-out
action sequences measures predictive accuracy independently of the
controller's gameplay return.

\paragraph{Executable models.}
Neural world models learn dynamics in latent or generative
form~\cite{ha2018world,hafner2025dreamer,bruce2024genie,valevski2025gamengen}.
WorldCoder, GIF-MCTS, Code World Models, and PoE-World synthesize or
repair programs over
supplied state interfaces~\cite{tang2024worldcoder,dainese2024gifmcts,lehrach2026cwm,piriyakulkij2025poeworld};
Mind-Studio uses OCAtari trajectories derived from random-access memory (RAM) in its reported
synthesis protocol~\cite{dong2026mindstudio}. OPINE-World, Tycho, and
Twin also revise observation abstractions, but study grid-game
settings~\cite{courtis2026opineworld,lehmann2026tycho,skoutnev2026twin}.
Visual program induction brings the input closer to pixels: Finite
Automata Extraction learns sprite programs without logged control
inputs~\cite{goel2026fae}, while VisualPatchWorld uses simulator-state
traces and dynamical sketch families~\cite{bai2026visualpatchworld}.
Code-as-World reconstructs physical scenes for quantitative
reasoning~\cite{wang2026codeasworld}, and D-World builds executable
robot-scene twins~\cite{wen2026dworld}. These methods draw on different
evidence, from supplied states and RAM trajectories to simulator traces
and visual observations. \method{} constructs both an Observer and a
stateful Predictor directly from visual gameplay, aligning executed
actions with tracked motion to build an executable transition model.

% !TEX root = ../main.tex
\section{Method}
\label{sec:method}

\method{} constructs a code world model (CWM) from recorded gameplay and
uses it to select actions during live play (Fig.~\ref{fig:method}).
The model combines a visual Observer $O$ and a stateful Predictor $P$
as defined in Eq.~\eqref{eq:observer-predictor}.

Table~\ref{tab:related-comparison} places \method{} among executable world
models. Its distinguishing combination is to build from raw visual gameplay
and recorded native actions, associate objects across frames, attribute their
continuous motion to executed actions, and compile observations
into a stateful Predictor. The Predictor supports counterfactual rollouts
for control and can be evaluated on held-out action sequences.
\begin{table}[t]
\centering% Source audit: table1_feature_audit.md. Marks describe compared configurations.
\begingroup
\fontsize{9.5}{11.5}\selectfont\setlength{\tabcolsep}{2pt}\renewcommand{\arraystretch}{1.14}
\begin{tabularx}{.96\columnwidth}{@{}l*{4}{>{\centering\arraybackslash}X}@{}}
\toprule
\raisebox{5pt}{Work} & \shortstack{Visual\\input} & \shortstack{Action\\input} & \shortstack{Cont.\\motion} & \shortstack{Auto\\attrib.}\\
\midrule
WorldCoder~\cite{tang2024worldcoder} & \no & \yes & \no & \no\\
GIF-MCTS~\cite{dainese2024gifmcts} & \no & \yes & \yes & \no\\
Code World Models~\cite{lehrach2026cwm} & \no & \yes & \no & \no\\
PoE-World~\cite{piriyakulkij2025poeworld} & \no & \yes & \yes & \no\\
Mind-Studio~\cite{dong2026mindstudio} & \no & \yes & \yes & \no\\
OPINE-World~\cite{courtis2026opineworld} & \no & \yes & \no & \no\\
Tycho~\cite{lehmann2026tycho} & \no & \yes & \no & \no\\
Twin~\cite{skoutnev2026twin} & \no & \yes & \no & \no\\
FAE~\cite{goel2026fae} & \yes & \no & \no & \no\\
Code-as-World~\cite{wang2026codeasworld} & \yes & \no & \yes & \no\\
\midrule
\textbf{\method} & \yes & \yes & \yes & \yes\\
\bottomrule
\end{tabularx}\endgroup

\caption{\textbf{Interfaces and capabilities in executable world modeling.}
Visual input: model induction from raw pixels without privileged state or
grid extraction; Action input: observed actions condition the learned model;
Cont. motion: non-grid coordinate dynamics; Auto attrib.: automatic separation
of action-driven from autonomous motion in those dynamics.
\yes: demonstrated in the reported configuration. \no: not demonstrated
under this definition; rendered game views alone do not qualify as visual input.
}
\label{tab:related-comparison}
\end{table}

% !TEX root = ../main.tex
\subsection{Problem Formulation}
\label{sec:formulation}
Let $s_{t}\in\mathcal S_{g}$ be the unobserved engine state of game $g$.
The stochastic transition kernel $\mathcal T_{g}$ maps a state, action,
and duration to a distribution over successor states; the rendering
function $\mathcal R_{g}$ maps an engine state to an image.
The environment evolves under a native action $a_{t}\in\mathcal A_{g}$ over an
executed interval $\Delta_{t}>0$, and renders an image $i_{t}$:
\begin{equation}
 s_{t+1}\sim \mathcal T_{g}(\cdot\mid s_{t},a_{t},\Delta_{t}),\qquad
 i_{t}=\mathcal R_{g}(s_{t}).
 \label{eq:environment}
\end{equation}
The construction system receives recorded gameplay images, executed native actions, action timing,
viewport metadata, and a declared textual description $d_{g}$.
When score $q_{t}$ and terminal/event signals $e_{t}$ are supplied by the
benchmark, they are explicit side information. Engine state and source code
are reserved for evaluation, not model construction.
Write $I_{t}=(i_{0},\ldots,i_{t})$ and $A_{t}=(a_{0},\ldots,a_{t})$ for the image
and action histories, with $A_{-1}$ empty. An episode ending after $T$ actions is
$\tau=(I_{T},A_{T-1},\Delta_{0:T-1},q_{0:T},e_{0:T})$.
Optional signals carry availability masks. Construction uses only the
observed prefix through $i_{t}$, including past actions and completed
intervals. Future actions are supplied separately as rollout inputs.

From these interactions, the system constructs a visual Observer $O$,
a stateful Predictor $P$ with numerical constants, and an optional
controller $\pi$, under a budget $B$ on calls, tokens, environment
interactions or wall time, as specified by each experiment.
The Observer maps the image sequence to a structured object history
$Y_{t}=O(I_{t})$, where $Y_{t}=(y_{0},\ldots,y_{t})$.
The Predictor uses the observation and action histories to predict the
next observation. Its inputs and internal components are specified below.
Observations $y_{t}$ from recorded prefixes
supply development supervision. Independent annotations $y_{t}^\star$
provide a reference for evaluating visual grounding and prediction.

The resulting model predicts observations under candidate actions, while its downstream
controller is evaluated in the real game:
\begin{equation}
 J_{g}(\pi)=\E_{\pi,\mathcal T_{g}}\!\left[\sum_{t=0}^{T-1}r_{t}\right],\qquad
 r_{t}=q_{t+1}-q_{t},
 \label{eq:return}
\end{equation}
when the declared objective is a cumulative score. Native win/loss utility
replaces this definition where appropriate. We assess prediction, return,
reliability, and construction cost as complementary outcomes.

\subsection{Observer Construction}
\label{sec:observer}
A VLM proposes entities to track from gameplay recordings and the
declared rules. GroundingDINO~\cite{liu2024groundingdino} and SAM~\cite{kirillov2023sam} supply candidate boxes and masks
on recorded frames. The VLM reviews the original frame and the indexed
candidate crops, then accepts a candidate or rejects them all. Accepted
annotations are pooled by entity class to train the local detector $D$.

A VLM-generated association program links detections across frames.
Its JSON context carries identities forward and resets at episode
boundaries. The program must be deterministic, preserve its inputs,
and use only observations available at the current step. Training and
development replay validate the contract, and errors are returned to
the VLM for revision. On each frame, the program uses the detections to
produce an associated observation $y_{t}$ and update its context $c_{t}$,
\begin{equation}
 (y_{t},c_{t})=U(D(i_{t}),c_{t-1}),
 \label{eq:observer}
\end{equation}
where $D$ detects entities and $U$ assigns persistent identities.
The context starts empty at each episode. Iterating this update yields
$Y_{t}=O(I_{t})$, a history of entity classes, boxes, and identities.
The supplementary material details annotation acceptance and the
association contract.

\subsection{Predictor Construction}
\label{sec:executable}
The overall Predictor maps the observed object history $Y_{t}$, action
history $A_{t}$ (including the candidate action $a_{t}$), and its duration
$\Delta_{t}$ to the next object observation:
\begin{equation}
 \widehat y_{t+1\mid t}=P(Y_{t},A_{t},\Delta_{t}).
 \label{eq:observer-predictor}
\end{equation}
The subscript $t+1\mid t$ marks a prediction anchored at the real prefix
through $t$.
The Predictor has three program components. First, \texttt{infer\_state}
initializes $z_{t\mid t}=\operatorname{Infer}_{P}(Y_{t},A_{t-1})$ from the real
Observer history and past actions. Second, \texttt{transition} advances
the state under a candidate action and duration,
$z_{t+1\mid t}=\operatorname{Transition}_{P}(z_{t\mid t},a_{t},\Delta_{t})$.
Third, \texttt{render} projects the updated state into screen-coordinate
objects, $\widehat y_{t+1\mid t}=\operatorname{Render}_{P}(z_{t+1\mid t})$.
Together, these components implement the overall prediction in
Eq.~\eqref{eq:observer-predictor}.
The controller repeats the transition and rendering for candidate action sequences, while
the latent state carries velocity, contact state, and timers across
simulated steps.

The underlying program operations preserve their inputs, and the state is
JSON serializable. Inference also receives recorded timing and available score
and event fields. Within each game, one sampling interval is the unit
of simulated time. Open-loop replay initializes the state once and
projects future observations into the anchor's screen coordinates.
Conditional one-step replay instead reinitializes from the real
Observer history at each anchor.

To construct the Predictor's transition rules, \method{} aligns
each executed action with the observed displacement of tracked objects.
For tracked object $j$ with screen-space box center $\mathbf p_{t}^j$,
action $a_{t}$ is paired with displacement
$\mathbf p_{t+1}^j-\mathbf p_{t}^j$. These action-conditioned displacement (ACD) records
also retain the object and association confidence. Comparing displacement
distributions across actions with a Kolmogorov--Smirnov (KS) distance and a
minimum effect threshold flags candidate input-dependent motion. The VLM
uses this evidence to distinguish action effects from autonomous dynamics
when writing the Predictor, and can query individual records for held-input
effects and discrete-event timing. The harness replays candidate programs
on development trajectories with recorded actions. Discrepancies between
predicted and observed objects then guide program revision, while
development-split performance determines which candidate is frozen for
test evaluation.

\subsection{Controller Construction}
\label{sec:control}
After construction, the VLM receives the Observer schema, the legal
actions, and the Predictor interface. It writes a Python function,
\texttt{choose\_action}, that generates candidate action sequences and
scores their Predictor-generated trajectories. We denote these two parts
of the controller $\pi$ by $\operatorname{Candidates}_{\pi}$ and
$\operatorname{Score}_{\pi}$ in Algorithm~\ref{alg:control}.
The function returns a native action and can retain context across calls.
The planning horizon $H$ is the number of simulated actions. For a candidate
sequence $\mathbf a$, let $v(\mathbf a)$ be its controller score.
\begin{algorithm}[t]
\caption{Game Control}
\label{alg:control}
\begin{algorithmic}[1]
\Require Observed histories $I_{t},A_{t-1}$; frozen $O,P$; controller $\pi$;
  horizon $H$ and candidate intervals $\Delta_{t:t+H-1}$
\State $Y_{t} \gets O(I_{t})$ \Comment{observe the current game state}
\State $z_{t\mid t}\gets\operatorname{Infer}_{P}(Y_{t},A_{t-1})$
\State $\mathcal C_{t} \gets \operatorname{Candidates}_{\pi}(Y_{t})$
\For{each $\mathbf a=(a_{t},\ldots,a_{t+H-1})\in\mathcal C_{t}$}
  \State $z\gets z_{t\mid t}$; $\widehat Y\gets[\,]$
  \For{$k=t,\ldots,t+H-1$}
    \State $z\gets\operatorname{Transition}_{P}(z,a_{k},\Delta_{k})$
    \State $\widehat y_{k+1\mid t}\gets\operatorname{Render}_{P}(z)$
    \State Append $\widehat y_{k+1\mid t}$ to $\widehat Y$
  \EndFor
  \State $v(\mathbf a)\gets\operatorname{Score}_{\pi}(\widehat Y)$
\EndFor
\State $\mathbf a^\star \gets \argmax_{\mathbf a\in\mathcal C_{t}} v(\mathbf a)$
\State Execute the first action of $\mathbf a^\star$
\end{algorithmic}
\end{algorithm}
For each candidate, $\operatorname{Transition}_{P}$ advances a separate copy of
the inferred state, and $\operatorname{Render}_{P}$ collects predicted observations without reading
future game frames. The controller scores the predicted sequence, executes
only the first action of the best candidate, and repeats the procedure
after the next real observation. The frozen $O$, $P$, and $\pi$ run without
online VLM calls; a runtime guard supplies a model-based fallback if $\pi$
fails or emits an invalid action.

% !TEX root = ../main.tex
\section{Experiments}
\label{sec:experiments}
% Declare the two full-width main-result tables early so they can appear
% at the top of the next page; their discussion remains in Sections 4.2--4.3.
% Raw provenance: control_table_data.json, with the updated DeepSeek suite.
\begin{table*}[t]
\centering\fontsize{9.5}{11.5}\selectfont\setlength{\tabcolsep}{2pt}
\caption{\textbf{\method{} raw control returns across model backbones and games.} Entries are mean native-game scores over ten evaluation episodes. Each measured model entry represents one available construction. Mean $Z$ is the equally weighted average of normalized returns across the eight games. In each column, the best model entry is \textbf{bold} and the runner-up \underline{underlined}. Random and Ref rows are calibration anchors, not ranked.}
\label{tab:control}
\begin{tabularx}{.90\textwidth}{@{}l*{9}{>{\centering\arraybackslash}X}@{}}
\toprule
\multirow{2}{*}{Backbone} & Chrome & Flappy & Doodle & Happy & \multirow{2}{*}{Asteroids} & Fruit & Road & Little & Mean \\
 & Dino & Bird & Jump & Hop & & Catcher & Fighter & Ski & $Z$ \\
\midrule
Random   &  58.2 &  0.0 &  100.8 &  0.0 &  440.0 &  30.6 &  165.5 &  46.2 &  0.000 \\
Ref   &  239.2 &  4.0 &  2584.9 &  17.6 &  6020.0 &  267.3 &  467.2 &  164.5 &  1.000 \\
\midrule
DeepSeek V4.1 Flash   &  101.6 &  0.1 &  265.4 &  1.5 &  2240.0 &  315.5 &  \underline{256.1} &  64.4 &  0.299 \\
Qwen3.8 Flash   &  45.2 &  0.0 &  316.2 &  1.7 &  70.0 &  \underline{393.1} &  149.0 &  74.1 &  0.220 \\
Kimi K3   &  46.5 &  0.0 &  534.1 &  0.0 &  2210.0 &  \textbf{469.1} &  165.8 &  193.4 &  0.441 \\
GLM-5.3-Flash   &  44.7 &  0.0 &  511.1 &  8.2 &  2090.0 &  327.9 &  157.5 &  \textbf{223.5} &  0.448 \\
MiniMax-M3   &  44.6 &  0.2 &  \underline{3018.5} &  \textbf{12.6} &  2640.0 &  15.7 &  114.0 &  123.3 &  0.335 \\
MiniMax-M3.1-Flash   &  182.0 &  0.0 &  37.4 &  1.7 &  1310.0 &  292.4 &  116.6 &  \underline{205.8} &  0.401 \\
GPT-6 Luna   &  \underline{191.4} &  \underline{1.2} &  719.3 &  \underline{9.3} &  \underline{2790.0} &  327.0 &  244.1 &  130.9 &  \underline{0.558} \\
GPT-6.1 Sol   &  \textbf{231.1} &  \textbf{2.3} &  \textbf{3672.8} &  0.0 &  \textbf{4160.0} &  122.7 &  \textbf{304.0} &  183.5 &  \textbf{0.706} \\
\bottomrule
\end{tabularx}
\end{table*}

% Provenance: Direct VLM (VLM-in-the-loop agent) and PlaySmith rows both use
% the deepseek-flash backbone, ten evaluation episodes per game. PlaySmith
% costs are one-time construction records (doodlejump from the 2026-11-08
% standard-protocol run).
\begin{table*}[t]
\centering\fontsize{9.5}{11.5}\selectfont\setlength{\tabcolsep}{4.5pt}
\caption{\textbf{Direct VLM agent versus \method{} under the same
DeepSeek V4.1 Flash backbone.} Native scores and cost over ten evaluation
episodes per game. The agent pays input tokens at every decision
step. \method{} spends a one-time construction budget and runs
evaluation without further model calls. Token counts are shown in
thousands for both.
Latency is measured in seconds per step during evaluation. The Mean row reports the eight-game mean normalized return $Z$ (marked $^{*}$), token costs, and per-step latency.}
\label{tab:direct-vlm-cost}
\begin{tabularx}{.90\textwidth}{@{}l*{8}{>{\centering\arraybackslash}X}@{}}
\toprule
\multirow{2}{*}{Game} & \multicolumn{4}{c}{Direct VLM agent} & \multicolumn{4}{c}{\method{}} \\
\cmidrule(lr){2-5}\cmidrule(lr){6-9}
& Score & Input (k) & Output (k) & Latency & Score & Input (k) & Output (k) & Latency \\
\midrule
Chrome Dino    & 61.4   & 3423 & 288.6 & 2.146  & 101.6   & 320  & 25.4 & 0.340 \\
Flappy Bird    & 0.0    & 1082 & 117.6 & 4.651  & 0.1     & 467  & 27.2 & 0.415 \\
Doodle Jump    & 230.9  & 996  & 191.7 & 5.251  & 265.4   & 1178 & 15.4 & 0.260 \\
Happy Hop      & 0.0    & 1215 & 97.0  & 12.714 & 1.5     & 603  & 24.7 & 0.366 \\
Asteroids      & 580.0  & 1648 & 189.9 & 3.633  & 2240.0  & 283  & 19.7 & 0.221 \\
Fruit Catcher  & 219.7  & 2701 & 225.3 & 1.052  & 315.5   & 531  & 27.0 & 0.236 \\
Road Fighter   & 196.2  & 883  & 100.4 & 2.730  & 256.1   & 366  & 24.3 & 0.298 \\
Little Ski     & 26.3   & 272  & 41.2  & 4.030  & 64.4    & 311  & 21.8 & 0.210 \\
\midrule
Mean$^{*}$     & 0.103 & 1527.5 & 156.5 & 4.526 & 0.299 & 507.4 & 23.2 & 0.293 \\
\bottomrule
\end{tabularx}
\end{table*}

\subsection{Experimental Setup}
\label{sec:setup}
We evaluate eight VLM backbones: DeepSeek-V4.1-Flash-1017
(DeepSeek V4.1 Flash in the tables)~\cite{deepseek1017},
Qwen3.8-Flash~\cite{qwen38flash}, Kimi K3~\cite{kimik3},
GLM-5.3-Flash~\cite{glm53flash}, MiniMax-M3~\cite{minimaxm3},
MiniMax-M3.1-Flash~\cite{minimaxm31}, GPT-6 Luna~\cite{gpt6luna},
and GPT-6.1 Sol~\cite{gpt61sol}.
\paragraph{Dataset.}
Our eight browser-based arcade games are Chrome Dino, Flappy Bird,
Doodle Jump, Happy Hop, Asteroids, Fruit Catcher, Road Fighter, and
Little Ski. We retain their native keyboard and pointer interfaces,
including actions that span multiple observations. Each build uses four
training and two development episodes, each capped at 400 decisions.
Control is evaluated on ten held-out episodes per game.

\paragraph{Implementation.}
Experiments run on an Ubuntu 22.04 server with 8 Intel Xeon vCPUs,
128 GiB RAM, and one NVIDIA A100 80 GB GPU.
The Observer uses GroundingDINO Swin-T OGC (box/text thresholds
0.15/0.10) and SAM ViT-B to propose and refine object annotations,
then trains a YOLOv8n~\cite{jocher2023yolov8} local detector from accepted crops (up to 30
epochs initially and 10 for warm starts). A VLM generates the
association program, Predictor, and controller. Construction uses one
round per build and freezes all three modules before test evaluation;
gameplay then runs without online VLM calls. The supplementary material
gives annotation filters, resource caps, and run manifests.

\paragraph{Evaluation metric.}

For each evaluated condition $c$ and game $g$, let $\bar R_{c,g}$ be the
mean raw return over ten held-out episodes, normalized as
$Z_{c,g}=(\bar R_{c,g}-\bar R_{\mathrm{random},g})/
(\bar R_{\mathrm{ref},g}-\bar R_{\mathrm{random},g})$.
Here Random is a weighted random policy over each game's native
action set, and Reference is a fixed privileged-state controller.
The two calibrate each game's score scale and are not bounds.

\subsection{How Does Gameplay Vary Across Backbones?}
Table~\ref{tab:control} compares eight backbones across the eight games.
GPT-6.1 Sol achieves the highest mean normalized return (0.706) and leads
on Chrome Dino, Flappy Bird, Doodle Jump, Asteroids, and Road Fighter,
though it scores zero on Happy Hop. GPT-6 Luna is second (0.558). The
remaining three game leads are distributed across three backbones: Kimi
K3 on Fruit Catcher, MiniMax-M3 on Happy Hop, and GLM-5.3-Flash on
Little Ski. This variation motivates reporting
both the equal-game mean and the individual game scores.

\subsection{Can \method{} Outperform Direct VLM Play at Lower Cost?}
We compare against a direct VLM agent that follows the GameWorld
agent-harness configuration~\cite{ouyang2026gameworld}: the game pauses
during inference, the model receives a screenshot with the game
description and native action list, reasons freely, may request up to
three zoomed crops for close inspection, and answers one native action
per step. The supplementary material details the full agent
configuration. Evaluation
seeds and episode caps are the same as for \method{}. Under the same
DeepSeek V4.1 Flash backbone, \method{} raises the
eight-game mean normalized return from 0.103 to 0.299
(Table~\ref{tab:direct-vlm-cost}), with higher mean raw scores in all eight games. The largest raw-score gain is on
Asteroids (580 to 2240), and Road Fighter rises from 196.2 to 256.1.
Per game, the direct agent uses 1,527.5k input tokens on average during
evaluation, versus 507.4k for \method{}'s one-time construction.
The resulting code plays all ten evaluation episodes per game
without further VLM calls. Mean per-step latency falls from 4.526 to
0.293 seconds across the eight games, with 67\% fewer input tokens per game
on average.
The compiled components can be reused for additional episodes.

\subsection{How Does the Predictor Compare with Other Construction Methods?}
\method{} has the highest horizon-3 geometric F1 among the compared
construction methods (Table~\ref{tab:construction-methods}).
We replace the \method{} Predictor-construction session with adapted
construction recipes inspired by other published systems and rebuild only the
Predictor, keeping the frozen Observer, the same DeepSeek V4.1 Flash
backbone, and the same 50k-token Predictor-construction budget. All
methods are evaluated frozen on the common test split. The cited papers
motivate the recipes; all numbers in Table~\ref{tab:construction-methods}
come from our adapted implementations, not the cited papers' experiments:
\begin{itemize}
\item \textbf{OPINE-World}~\cite{courtis2026opineworld}: counterexample-guided
whole-module rewriting, a Bayesian evidence ledger, critic review, and
live-game experiments.
\item \textbf{DreamTeam}~\cite{sarafian2026dreamteam}: six roles
(Observer, Simulator, Inductive Explorer, Transductive Explorer,
Critic, and Team Leader) with
commit-then-retrodict and regression-set replay.
\item \textbf{Continual}~\cite{karten2026continual}: starting from the
minimal persistence seed, a Refiner reads unstructured monotone failure
logs after each evidence round and emits function-level incremental
patches.
\item \textbf{OpenCode}~\cite{opencode2025}: the open-source terminal
coding agent, given the same evidence, tools and budget through its
build/plan agent pair.
\item \textbf{Persistence floor}: the copy-last-frame persistence baseline,
scored on the same split as references.
\end{itemize}
The primary metric is entity-level geometric F1 at horizon 1 ($h_{1}$).
For each window and entity class, predicted boxes are paired with
recorded boxes by maximum-cardinality one-to-one matching at an
intersection-over-union (IoU) threshold of $0.5$. Matched pairs count as true positives, unmatched
predicted boxes as false positives, and unmatched recorded boxes as
false negatives, so misplacements and object-count mismatches are
both penalized. Scores are micro-averaged over all windows
and classes, then macro-averaged over the eight games.

\begin{table}[t]
\centering\fontsize{9.5}{11.5}\selectfont\setlength{\tabcolsep}{2pt}
\caption{\textbf{Predictor-construction method comparison.} Eight-game macro-average geometric F1 at horizons 1 and 3 ($h_{1}$, $h_{3}$), normalized return $Z$, and per-game construction cost. All learned methods use DeepSeek V4.1 Flash; each Predictor is paired with the same frozen Observer and controller.}
\label{tab:construction-methods}
\begin{tabularx}{.95\columnwidth}{@{}>{\raggedright\arraybackslash}p{.41\columnwidth}*{4}{>{\centering\arraybackslash}X}@{}}
\toprule
Construction & $h_{1}$ & $h_{3}$ & $Z$ & Tokens\\
\midrule
Continual & 0.682 & 0.500 & 0.234 & 49k\\
OPINE-World
 & 0.590 & 0.417 & 0.218 & 50k\\
DreamTeam & 0.508 & 0.396 & 0.177 & 47k\\
OpenCode & 0.465 & 0.359 & 0.172 & 49k\\
Persistence floor & 0.304 & 0.089 & 0.109 & 0\\
\midrule
\textbf{Ours} & \textbf{0.835} & \textbf{0.694} & \textbf{0.299} & 50k\\
\bottomrule
\end{tabularx}
\end{table}

Table~\ref{tab:construction-methods} reports $h_{1}$, $h_{3}$, normalized
return $Z$, and per-game Predictor-construction cost. The horizon-3
degradation is milder for \method{} than for any baseline (a 17\%
relative drop, 0.835 to 0.694, versus 22--29\%), so the prediction
advantage persists across rollout horizons. $Z$ isolates the Predictor's
contribution to control: the same frozen Observer and the same frozen
controller are paired with each method's Predictor, so differences in
$Z$ reflect prediction quality alone.

Successive errors distinguish Predictors with similar one-step accuracy.
Table~\ref{tab:construction-methods} measures geometric F1 at
individual offsets. We also replay recorded actions from each test
anchor and measure $S(k)$, the fraction of rollouts whose object
geometry and counts satisfy the joint correctness criterion at every
step through $k$. The Predictor initializes from the observed history
at the anchor and then advances its simulated state without seeing
later frames. In Flappy Bird, \method{} and the strongest alternative
have the same one-step survival rate, but at five steps they retain
44\% and 26\% of anchors, respectively. The gap shows how successive
errors can distinguish Predictors with similar immediate accuracy.
Four of the eight games reach the floor before step five for all
construction methods. The supplementary material reports the full
per-game curves.

\subsection{How Accurate Is the Observer?}
% Provenance: validation/observer_eval/by_model_summary_v4.json — replayed
% action trajectories with frozen detectors on the same pixels; game-internal
% state as ground truth; eight-game unweighted means. Element names are mapped
% to the game vocabulary by alias matching before scoring.
\begin{table}[t]
\centering\fontsize{9.5}{11.5}\selectfont
\caption{\textbf{Observer accuracy across VLM backbones.} Each backbone
constructs its own entity vocabulary, detector, and association code.
Detection F1 counts a pair when IoU $\geq 0.5$. Matched IoU (mIoU) averages IoU over
matched pairs. Mean center distance (CD) is in viewport pixels. Values are
unweighted means over the eight games.}
\label{tab:observer-main}
\setlength{\tabcolsep}{1pt}
\fontsize{8.5}{10.5}\selectfont
\begin{tabularx}{.95\columnwidth}{@{}l*{5}{>{\centering\arraybackslash}X}@{}}
\toprule
Backbone & Prec. & Recall & F1 & mIoU & CD\\
\midrule
DeepSeek V4.1 Flash & 0.938 & 0.924 & 0.925 & 0.73 & 7.0\\
Kimi K3 & 0.949 & 0.892 & 0.916 & 0.72 & 5.6\\
Qwen3.8 Flash & 0.938 & 0.896 & 0.909 & 0.71 & 11.6\\
MiniMax-M3.1-Flash & 0.907 & 0.890 & 0.892 & 0.71 & 7.4\\
GLM-5.3-Flash & 0.884 & 0.903 & 0.889 & 0.67 & 12.6\\
GPT-6 Luna & 0.891 & 0.890 & 0.885 & 0.62 & 13.3\\
MiniMax-M3 & 0.813 & 0.870 & 0.825 & 0.64 & 19.9\\
GPT-6.1 Sol & 0.753 & 0.717 & 0.733 & 0.69 & 10.2\\
\bottomrule
\end{tabularx}
\end{table}

Table~\ref{tab:observer-main} scores the
Observer by replaying each backbone's recorded action trajectory and
reading detector output and engine-exported boxes on the same frozen
frames. Their matched-IoU values
fall within a narrow band of 0.62--0.73: Observer quality is dominated by the
task rather than the backbone. GPT-6.1 Sol is the exception ($F_{1}=0.733$,
recall $0.717$), yet it achieves the highest control return in
Table~\ref{tab:control}. Control return and Observer F1 rank backbones
inversely (Spearman $\rho=-0.62$ over eight backbones,
not significant at $n=8$): more accurate
perception does not visibly translate into better gameplay, and the gap
between models concentrates in dynamics prediction rather than
perception quality. Failure modes are systematic rather than random: six of eight
backbones are recall-limited, with misses concentrated on small or rare
entities (bullets, springs, moving platforms) while large objects such as
pipes and vehicles reach matched IoU above 0.8.

% !TEX root = ../main.tex
\section{Ablation Studies}
\label{sec:ablation}
\subsection{Does the Predictor Improve Control?}
We disable access to the Predictor at control time and compare the
resulting reactive-only policy with the full system.
Table~\ref{tab:reactive_only} shows higher native-game returns with
Predictor access in all eight games.

\begin{table}[t]
\centering\fontsize{9.5}{11.5}\selectfont\setlength{\tabcolsep}{4.5pt}
\caption{\textbf{Predictor ablation.} Returns are in native game units.
w/o Pred. disables Predictor access during control; Diff. is w/o Pred. minus Full.}
\label{tab:reactive_only}
\begin{tabularx}{.88\columnwidth}{@{}l*{3}{>{\centering\arraybackslash}X}@{}}
\toprule
Game & Full & w/o Pred. & Diff.\\
\midrule
Fruit Catcher  & 315.5   & 115.0 & $-200.5$\\
Chrome Dino    & 101.6   & 48.3 & $-53.3$\\
Doodle Jump    & 265.4  & 151.3 & $-114.1$\\
Happy Hop      & 1.5     & 0.0 & $-1.5$\\
Flappy Bird    & 0.1     & 0.0 & $-0.1$\\
Asteroids      & 2240.0   & 1880.0 & $-360.0$\\
Road Fighter   & 256.1   & 156.6 & $-99.5$\\
Little Ski     & 64.4    & 20.3 & $-44.1$\\
\bottomrule
\end{tabularx}
\end{table}

Normalized by the Reference--Random gap, the largest improvement occurs
in Fruit Catcher (0.847), followed by Little Ski (0.373) and Road
Fighter (0.330). In contrast, Asteroids achieves the largest absolute
gain (360 points), but only a 0.065 normalized improvement.

\subsection{Why Does Our Predictor Work?}
We vary the evidence and feedback available to DeepSeek V4.1 Flash during Predictor
construction while preserving the recorded splits, synthesis loop and
budget. Test replay uses recorded actions.
Table~\ref{tab:ablation} reports the four arms.
Removing the ACD records is the most damaging arm: $h_{1}$ drops from
0.835 to 0.536 and $Z$ from 0.299 to 0.193, so the aligned
action-motion pairs are the main source of construction signal.
Shuffled action labels recover less than a third of this gap ($h_{1}$
0.631, $Z$ 0.204): correct attribution, not record volume, carries the
gain, and label integrity matters mostly for attribution rather than
for the diagnostic content of feedback. Removing the KS attribution
test ($h_{1}$ 0.702, $Z$ 0.250) removes a substantial source of
construction signal: the verdict is a
revisable hypothesis, yet the VLM reconstructs only part of the action
grouping from the remaining ACD records and replay feedback, and its
lead over the strongest external baseline narrows from 0.153 to 0.020.
Dropping the replay-feedback loop ($h_{1}$ 0.782, $Z$ 0.241)
carries the mildest $h_{1}$ cost. The four arms rank the evidence: the
paired records provide the main signal, correct attribution anchors it, the
statistical verdict contributes a margin, and the feedback loop
yields the smallest refinement.

\begin{table}[t]
\centering\fontsize{9.5}{11.5}\selectfont\setlength{\tabcolsep}{2pt}
\caption{\textbf{Predictor construction ablation.} 8-game macro-average
geometric F1 at horizons 1 and 3 ($h_{1}$, $h_{3}$), normalized return
$Z$, and per-game construction cost. Conditions use DeepSeek V4.1 Flash and the same frozen
Observer and controller.}
\label{tab:ablation}
\begin{tabularx}{.95\columnwidth}{@{}>{\raggedright\arraybackslash}p{.41\columnwidth}*{4}{>{\centering\arraybackslash}X}@{}}
\toprule
Construction & $h_{1}$ & $h_{3}$ & $Z$ & Tokens\\
\midrule
\textbf{Full} & \textbf{0.835} & \textbf{0.694} & \textbf{0.299} & 50k\\
w/o KS attribution & 0.702 & 0.577 & 0.250 & 46k\\
w/o ACD records & 0.536 & 0.363 & 0.193 & 48k\\
shuffled actions & 0.631 & 0.483 & 0.204 & 49k\\
w/o feedback & 0.782 & 0.653 & 0.241 & 48k\\
\bottomrule
\end{tabularx}
\end{table}

% !TEX root = ../main.tex
\section{Case Analysis}
\label{sec:case-analysis}
\begin{figure*}[t]
\centering
\includegraphics[width=\textwidth]{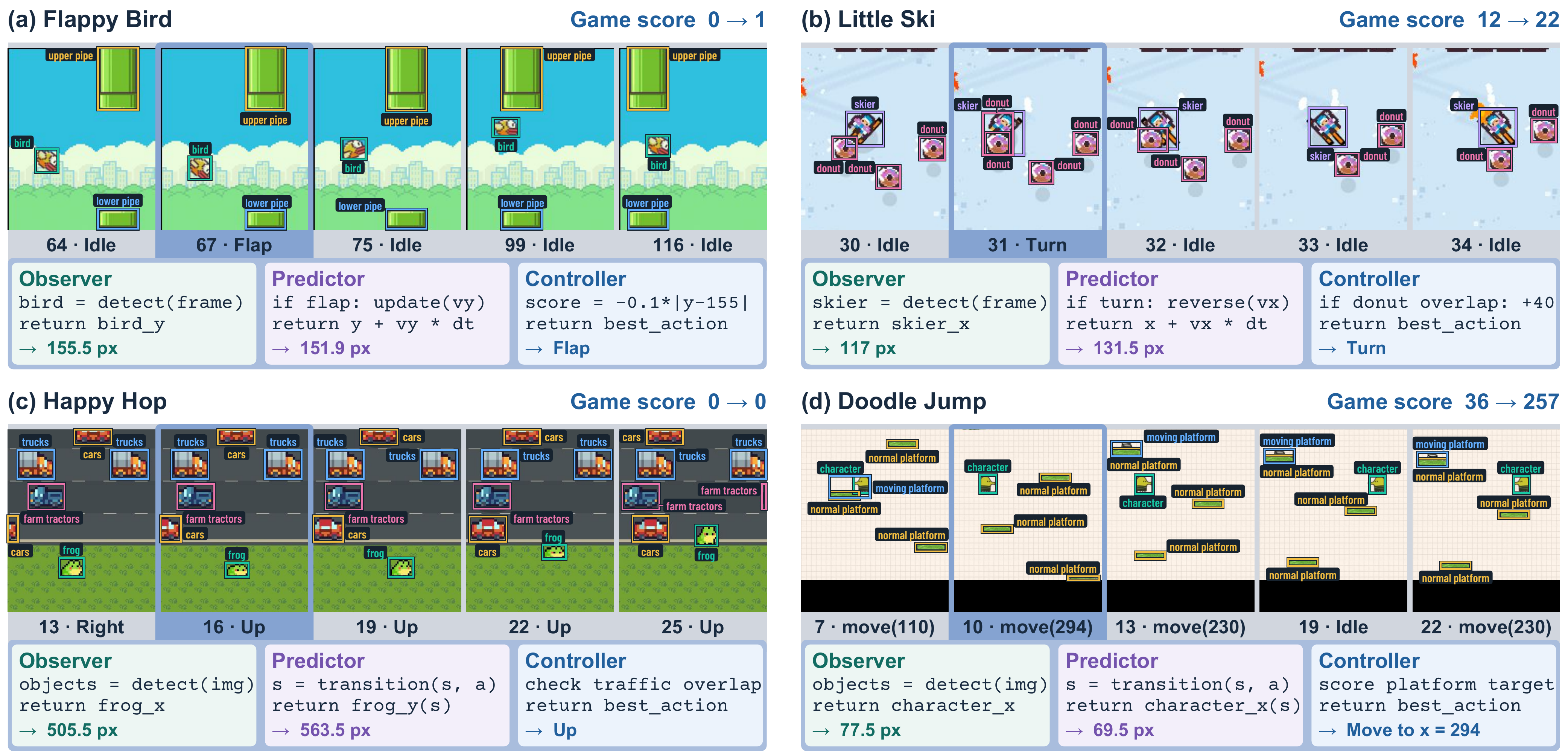}
\caption{\textbf{Closed-loop control in four games.}
Each strip shows five sampled frames from a real evaluation episode, with
Observer detections overlaid. Top to bottom: Flappy Bird, Little Ski,
Happy Hop, and Doodle Jump. In each strip, a shared background groups the second frame with its three pseudocode panels. The panels show how the Observer
extracts object state, the Predictor advances it, and the controller
scores candidate actions. The scores above the strips are in-game scores.}
\label{fig:gameplay-control}
\end{figure*}

Fig.~\ref{fig:gameplay-control} traces one live decision in each
of four games. The highlighted frame supplies the observation,
the Predictor advances the state under candidate inputs, and the
controller selects an action using the predicted outcomes.

In Little Ski, the Observer estimates the skier's horizontal position
at 117 pixels. Reversing the movement direction at step 31 yields a
predicted position of 131.5 pixels. This lateral shift changes the
skier's projected overlap with a donut, which the controller rewards.
The selected turn is followed by an increase from 12 to 22 in the
sampled game sequence. In Doodle Jump, the controller uses predicted
character motion to score platform targets. At the highlighted step,
the observed horizontal position is 77.5 pixels, and the Predictor
returns 69.5 pixels for the illustrated update. The selected
\texttt{move(294)} specifies an input target in the game interface.
The predicted character position and the input target serve different
roles in the decision.

The other two cases illustrate how the scoring rule changes with the
game. Flappy Bird scores the predicted bird height against a target
height, while Happy Hop checks predicted overlap with traffic. These
examples show what information the controller can use when prediction
is enabled. Table~\ref{tab:reactive_only} measures the effect of
Predictor access across the evaluation episodes.

%\FloatBarrier  % removed: caused premature column break (empty right column on p8)
% !TEX root = ../main.tex
\section{Conclusion}
\label{sec:conclusion}
We presented \method{}, a code world model built from visual gameplay.
Its Observer tracks objects with persistent identity, while ACD evidence
guides a stateful Predictor. The frozen model supports multi-step
simulation and control without online VLM calls. Across eight games,
\method{} outperforms a direct VLM agent on the shared backbone while
using one-third as many aggregate input tokens. Observer quality varies
mainly by game rather than backbone, and the compiled system reduces
aggregate workflow time across ten episodes per game. Its prediction advantage
persists across rollout horizons, demonstrating the value of inspectable,
reusable dynamics.

% !TEX root = ../main.tex
\section*{LLM Usage Statement}
OpenAI Codex assisted with research ideation, related-work organization,
mathematical formulation, experimental design, manuscript drafting and editing,
LaTeX, tables, and programmatic figures. The PlaySmith pipeline uses the
VLM backbone listed in each results table for visual entity identification
and the generation and revision of Observer, dynamics, and controller
programs. Human authors remain responsible for the
claims, citations, code, and results reported in the manuscript.

{\small\bibliography{main}}

% =====================================================================
% Supplementary material (kept from the CVPR submission's appendix)
% =====================================================================
\clearpage
\appendix
% !TEX root = ../supplementary.tex
\clearpage
\appendix
\section*{Supplementary Material}
This supplement accompanies the main paper in the same document. It specifies
construction interfaces, objectives, controlled comparisons, and diagnostics.

\section{Visual Observer and Annotation Acceptance}
\label{app:observer}
\paragraph{Grounding and masks.}
The grounding configuration is GroundingDINO Swin-T OGC
(\texttt{groundingdino\_swint\_ogc.pth}), with box/text thresholds
0.15/0.10 and at most five candidates. Boxes occupying more than half the
frame are discarded before candidate selection. SAM ViT-B
(\texttt{sam\_vit\_b\_01ec64.pth}) refines each candidate. The VLM sees
the native frame and indexed mask crops, then returns a candidate index or
rejection. An invalid index, empty mask, nonfinite geometry, or box outside
the viewport does not create a training label. Zoomed appearance anchors
must also contain foreground evidence: channel standard deviation at least
6, or brightness differing by at least 25 from a background mode accounting
for at least half the crop. Thresholds operate on 8-bit pixels.
These rules filter malformed or background-only evidence; they do not prove
semantic correctness. All candidates and rejections are retained for audit.

\paragraph{Local detector.}
Accepted training crops are pooled by entity class. We use a YOLOv8n detector
initialized from \texttt{yolov8n.pt}, with a 30-epoch cold-start cap;
subsequent rounds warm-start for at most 10 epochs only when the complete
class-index mapping matches. A changed class list forces a new detector.
Training and validation image partitions follow complete episodes, with no
held-out test labels or frames. The local execution condition uses local
text extraction where required and makes no online VLM requests. Detector
selection uses only development evidence. Full training arguments, image
manifests, and weight digests belong to the released run manifest.

\paragraph{Automatic versus human operations.}
Entity proposals, mask choice, detector construction, association code, and
coordinate code are generated or selected automatically within the declared
budget. Human work sets the benchmark's controls, resets, and evaluation
labels; it does not correct training pseudo-labels in the automatic condition.
Manually revised pilot artifacts are excluded from that condition. Independent
evaluation annotators do not see model predictions when creating labels.

\paragraph{Association.}
The association program is VLM-generated Python, not a universal learned
tracker. It consumes detections, the previous JSON context, and past timing
and action metadata. It returns associated objects and new context.
Validation executes every
training and development frame, checks determinism, JSON serialization and
input preservation, and resets context at episode boundaries. No current
observation may read a future image. Identity labels used for evaluation are
never supplied to these programs.

\paragraph{Freezing.}
The study freezes perception before comparing dynamics models. A failed or
unsupported proposed entity remains unresolved rather than being repaired by
a human. Observer construction and failed proposals are charged to the same
development resource account.

\paragraph{Measured annotation accuracy.}
\label{app:observer-accuracy}
Observer boxes are scored against each game's internal state: the recorded
action trajectory is replayed under the runner's freeze contract, and at
every step the frozen detector runs on the captured frame while ground-truth
boxes are read from the game API and projected to viewport pixels. Pairing
both reads at the same frozen instant isolates detector quality from
replay and timing divergence; a two-snapshot bracket absorbs the
sub-tick phase between the state update and the canvas draw. Element names
are mapped to the game vocabulary by alias matching before scoring, so
per-backbone naming choices do not count as misses. Table~\ref{tab:observer-main}
summarizes the outcome; Table~\ref{tab:observer-per-game} gives the full
per-game matrix.

The per-game matrix shows three structural effects that are independent of
the backbone. First, difficulty is dominated by object size: large on-screen
objects such as road vehicles and pipes reach F1 above 0.85 for every
backbone, whereas the doodlejump and littleski matched-IoU means are 0.41
and 0.37. In those two games the failure is mostly box-scale shrinkage
rather than mislocalization: matched center distances (8--15\,px) are
comparable to the other games while the boxes are systematically smaller
than the engine-defined ground-truth rectangles.\footnote{The ground truth
uses each game's engine rectangles (collision or sprite bounds), which can
differ from the visible sprite extent the detector is trained on; the IoU
gap therefore partly reflects box-convention mismatch rather than
localization error.} Second, rare categories fail in two opposite
directions: low-frequency entities such as springs and moving platforms are
either never registered (all ground-truth boxes missed, e.g.\ 117 spring
boxes for DeepSeek V4.1 Flash) or over-generalized (MiniMax-M3 emits about 180
moving-platform boxes per frame). Third, the bullet class is prone to
over-firing under the YOLO cold start on tiny objects: Kimi K3 and
MiniMax-M3 produce 405 and 144 false positives respectively against a
$6\times6$\,px target on a $600\times500$ canvas.

Backbone differences are more informative in stability than in mean: the
per-game F1 standard deviation of MiniMax-M3 (0.158) is three times that of
DeepSeek V4.1 Flash (0.053), and its worst game (doodlejump, 0.54) sits 0.45
below its best, while Kimi K3 and Qwen3.8 Flash each reach $F_{1}\geq 0.9$
on six of eight games. For downstream use, a high-variance Observer implies
unpredictable perception quality, which is harder to compensate than a
uniformly lower mean.

% Provenance: validation/observer_eval/by_model_summary_v4.json — same
% replay/same-frame protocol as the main-text Observer table; per-game cells
% are detection F1 (IoU>=0.5) and matched-pair IoU.
\begin{table*}[t]
\centering\fontsize{9.5}{11.5}\selectfont\setlength{\tabcolsep}{3.2pt}
\caption{\textbf{Observer accuracy per game and backbone.} Cells report
detection F1 (IoU $\geq$ 0.5) / matched-pair IoU against game-internal
ground truth on replayed trajectories, two episodes per game. Element
categories the Observer never registered count as misses for recall;
Observer categories with no ground-truth counterpart count as false
positives. Game names follow Table~\ref{tab:control}.}
\label{tab:observer-per-game}
\begin{tabular}{lr@{\;/\;}r@{\;/\;}r@{\;/\;}r@{\;/\;}r@{\;/\;}r@{\;/\;}r@{\;/\;}r@{\;/\;}r}
\toprule
\multirow{2}{*}{Backbone} & \shortstack{Chrome\\Dino} & \shortstack{Flappy\\Bird} & \shortstack{Doodle\\Jump} & \shortstack{Happy\\Hop} & \shortstack{Astero\\ids} & \shortstack{Fruit\\Catcher} & \shortstack{Road\\Fighter} & \shortstack{Little\\Ski} & \multirow{2}{*}{\shortstack{Mean\\F1}}\\
& \multicolumn{8}{c}{F1\,/\,matched IoU} & \\
\midrule
DeepSeek V4.1 Flash & 1.00\,/\,0.64 & 0.84\,/\,0.91 & 0.88\,/\,0.60 & 0.90\,/\,0.84 & 0.92\,/\,0.82 & 0.96\,/\,0.89 & 1.00\,/\,0.76 & 0.90\,/\,0.41 & \textbf{0.925}\\
Kimi K3 & 1.00\,/\,0.65 & 0.99\,/\,0.92 & 0.90\,/\,0.51 & 0.93\,/\,0.88 & 0.76\,/\,0.82 & 0.96\,/\,0.88 & 0.98\,/\,0.77 & 0.81\,/\,0.34 & 0.916\\
Qwen3.8 Flash & 0.90\,/\,0.74 & 0.77\,/\,0.74 & \textbf{0.91}\,/\,0.46 & 0.93\,/\,0.89 & 0.93\,/\,0.82 & 0.95\,/\,\textbf{0.91} & 0.98\,/\,0.67 & \textbf{0.90}\,/\,0.42 & 0.909\\
MiniMax-M3.1-Flash & 1.00\,/\,0.65 & 0.99\,/\,0.91 & 0.75\,/\,0.52 & 0.90\,/\,0.89 & 0.86\,/\,0.81 & 0.94\,/\,0.87 & 0.96\,/\,0.68 & 0.73\,/\,0.34 & 0.892\\
GLM-5.3-Flash & 1.00\,/\,0.65 & 0.98\,/\,\textbf{0.92} & 0.72\,/\,0.21 & 0.86\,/\,0.88 & 0.76\,/\,0.82 & 0.96\,/\,0.87 & \textbf{0.99}\,/\,0.67 & 0.84\,/\,0.37 & 0.889\\
GPT-6 Luna & 0.78\,/\,0.66 & 0.82\,/\,\textbf{0.92} & 0.87\,/\,0.22 & \textbf{0.95}\,/\,\textbf{0.88} & \textbf{0.94}\,/\,0.81 & 0.95\,/\,0.48 & 0.98\,/\,0.67 & 0.79\,/\,0.35 & 0.885\\
MiniMax-M3 & 0.99\,/\,0.66 & 0.85\,/\,0.60 & 0.54\,/\,0.33 & 0.93\,/\,0.85 & 0.88\,/\,\textbf{0.83} & 0.60\,/\,0.85 & 0.98\,/\,0.64 & 0.84\,/\,\textbf{0.38} & 0.825\\
GPT-6.1 Sol & 0.72\,/\,0.65 & \textbf{0.99}\,/\,0.92 & 0.34\,/\,0.35 & 0.86\,/\,0.87 & 0.92\,/\,0.82 & 0.94\,/\,0.90 & 0.91\,/\,0.68 & 0.17\,/\,0.33 & 0.733\\
\bottomrule
\end{tabular}
\end{table*}

\section{Visual Evidence in Scrolling Scenes}
\label{app:visualscroll}
Horizontal scene motion in Chrome Dino and vertical scene motion in Doodle
are examples of changing screen observations within the evaluated domain.
Screen-space Predictors represent such motion through object displacements.
The projection stays anchored to the observation frame, whereas
the executable state and projected object geometry continue to evolve.

\begin{figure}[t]
\centering
\includegraphics[width=\linewidth]{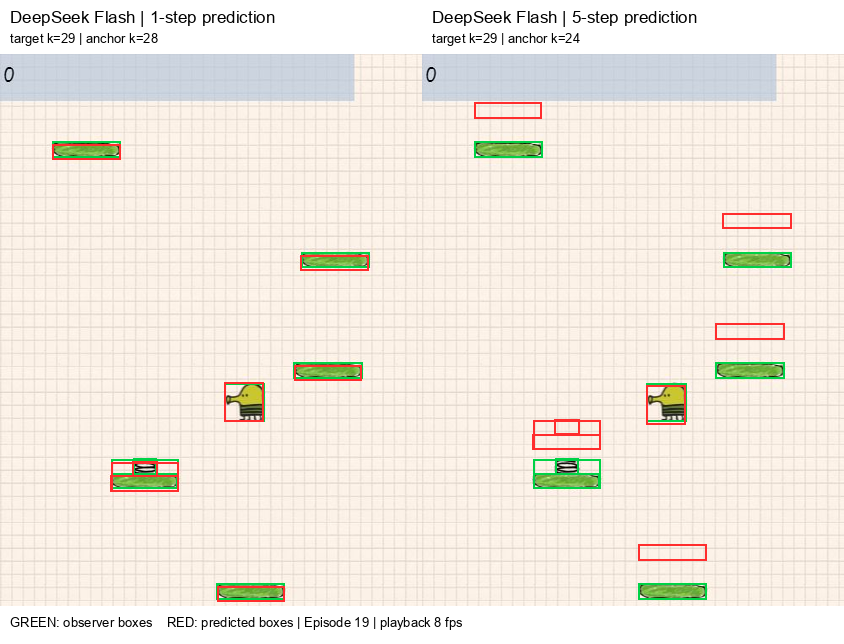}
\caption{\textbf{Existing Doodle pilot in a scrolling scene.} Left: one-step
prediction from anchor $k=28$; right: five-step prediction from anchor $k=24$.
Both target $k=29$ of episode 19. Green: target Observer boxes. Red: predicted
boxes. The longer rollout exposes misaligned and unmatched predictions.
This is a qualitative pilot illustration, not a suite-level result.}
\label{fig:scrollpilot}
\end{figure}

Fig.~\ref{fig:scrollpilot} reproduces an existing DeepSeek V4.1 Flash Doodle
pilot preview, distinct from the strong-backbone comparison and suite means.
Both panels show the same target frame, with predictions anchored one or
five sampling steps earlier. Green boxes are the Observer outputs at the
target frame, not independent manual labels; red boxes are predictions.
Several platforms align closely at one step, while five-step predictions
show displacement and object-count mismatches. The overlay illustrates how
geometric agreement changes with the prediction horizon. Quantitative
prediction and identity metrics use the separate evaluation protocols.

\section{Development Scoring}
\label{app:objectives}
\paragraph{ACD divergence test.}
For entity $\ell$ and action $a$, let $V_{\ell,a}$ collect the
per-transition signed displacement rates of $\ell$ under $a$, in pixels
per sampling step, computed from mutually associated boxes (tracked
identities take precedence; untracked mutual-nearest associations are
marked as hypotheses), and let $\widehat F_{\ell,a}$ denote the empirical
cumulative distribution function (CDF) of $V_{\ell,a}$. An axis ($dx$ or $dy$) of $\ell$ is flagged as
action-conditioned only when every supported pair of actions $a,a'$
(each with $|V_{\ell,a}|\geq 20$ samples) satisfies both
\begin{equation}
 \begin{aligned}
 \sup_{v}\bigl|\widehat F_{\ell,a}(v)-\widehat F_{\ell,a'}(v)\bigr|
 &\geq \max\bigl(0.25,\;\eta_{\ell,a,a'}\bigr),\\
 \eta_{\ell,a,a'}
 &=\sqrt{\frac{\ln(4m/0.01)}{2}
 \left(\frac{1}{|V_{\ell,a}|}+\frac{1}{|V_{\ell,a'}|}\right)}.
 \end{aligned}
 \label{eq:acddivergence}
\end{equation}
and a mean absolute quantile difference of at least $1$ pixel per
sampling step between the two samples (quantiles at nineteen evenly
spaced levels). Here $m$ is the number of supported action pairs compared
within a single entity-axis test; the logarithmic constant is the
implemented conservative Bonferroni adjustment, holding the KS
family-wise level at $0.01$ across the $m$ pairwise tests of that one
test (per-pair level $0.01/2m$, twice the textbook $0.01/m$), floored at
a distance of $0.25$. Entities with no flagged axis are treated as
action-invariant background; untracked association provenance is
retained with every pair.

\paragraph{Selection objective.}
Let $n_{h},w_{h}$ denote eligible windows and wrong predictions for $h=1,3$.
Eligibility for conditional accuracy requires nonempty equal-count frames;
correctness requires a complete class-consistent one-to-one IoU-$0.5$ match.
The report separately records excluded empty windows, count mismatches, and
coverage. The development loss adds the evaluator's structural-count penalty
$C$ to the one-step errors, capped at $n_{1}$:
\begin{equation}
 \bar w_{1}=\min(n_{1},w_{1}+C),\quad \bar w_{3}=w_{3}.
\end{equation}
$C$ accumulates reported count mismatches beyond the observed birth/death
allowance over evaluated rollout rows. Count differences equal to newly
appearing or disappearing objects are logged as spawn-noise exclusions.
This is a heuristic training allowance, not oracle identification of
unpredictable births. The test F1 counts all missed and extra objects.
Define the integer excess
\begin{equation}
 E=\sum_{h\in\{1,3\}}\max\!\left(0,
 \bar w_{h}-\left\lfloor\frac{\alpha_{h} n_{h}-1}{100}\right\rfloor\right),
\end{equation}
where $\alpha_{1}=20,\alpha_{3}=30$, omitting an unavailable $h=3$ row.
Then the development selection loss is
\begin{equation}
 \mathcal L_{\mathrm{sel}}=E(n_{1}+n_{3}+1)+\bar w_{1}+\bar w_{3}+
 \frac{A}{1+A}.
 \label{eq:actualsel}
\end{equation}
$A\geq0$ is the auxiliary diagnostic residual with weights
$(0.5,1,0.5,1,0.1,0.2,0.5,0.5)$ for one-step strict error,
persistent-object geometry, IoU error, size error, presence, score, event,
and multi-step error respectively. Presence is capped at 0.5 before weighting.
The bounded final term breaks integer-error ties; it cannot compensate for
more primary errors. Invalid execution or missing mandatory evidence yields
infinite selection loss. Ties retain the earliest submitted version.
Development split evidence is required for the formal study; exploratory
training-only fallback is not used for reported model selection.

\paragraph{Optional numerical facility.}
The repository also supports train-only bounded calibration of declared
\texttt{PARAMS}. It serves as an auxiliary procedure alongside ACD-informed
construction. It uses fixed comparison targets, an accuracy-prioritized
geometry objective and bounded coordinate search; missing geometry or runtime
errors cannot produce a zero loss. Candidate declarations and search traces
identify runs that explicitly enable it. It is an implementation option,
not a claimed contribution or an assumed explanation of the reported gains.

\section{Prompt, Tool, and Candidate Contracts}
\label{app:interfaces}
The constructor receives the operational description, observation schema,
coordinate units, available training/development episode references, and
resource limits. Its central instruction is:
\begin{quote}\small
Write Python defining infer\_state(history), transition(state, action, dt),
and render(state). Preserve all inputs. Store persistent latent information
in JSON-serializable state. Only transition advances simulated time. render
is read-only state projection. Access
only the supplied observed prefix; do not read game source, engine variables,
held-out labels, or test episodes. Declare numerical parameters and their
units. Use training evidence for numerical analysis, and development
execution evidence for candidate selection. Submit complete source after each repair.
\end{quote}

\begin{table}[t]
\centering\small
\caption{Construction-loop operations and evidence boundaries.}
\label{tab:tools}
\begin{tabularx}{\columnwidth}{lX}
\toprule
Operation & Required behavior\\
\midrule
Inspect & Read a named train/dev frame range and overlays; no test paths.\\
Analyze & Execute bounded numeric analysis over supplied train references.\\
Submit & Validate source; assign version and source digest.\\
Evaluate & Unroll specified anchors with recorded actions and intervals.\\
Fit & Change declared constants on training windows only.\\
Compare & Score versioned candidates on common dev windows.\\
Finish & Freeze the best valid dev candidate and resource totals.\\
\bottomrule
\end{tabularx}
\end{table}
Table~\ref{tab:tools} summarizes the construction-loop operations and their
evidence boundaries. A tool result carries success/failure, version, source digest,
eligible counts, error traces, and remaining budget. Failed proposals remain
in the transcript and cost totals. Repair feedback includes a failing anchor,
visible history, action sequence, durations, predicted latent state, projected
objects, and reference objects. Such feedback does not automatically label
an error as perceptual, structural, or numerical. Only training/development
paths are exposed to code execution. The frozen runtime checks input mutation,
serialization, repeated-call determinism, and valid projected output fields.

\section{Evaluation Manifests and Evidence Boundaries}
\label{app:manifest}
The primary panel contains Chrome Dino, Flappy Bird, Doodle Jump, Happy Hop,
Asteroids, Fruit Catcher, Road Fighter, and Little Ski. Each run manifest
records game and adapter digests, viewport, reset procedure, controls,
hold/pulse semantics, sampling intervals, episode cap, objective, terminal
rules, and the verbatim text supplied to every constructor. Numerical dynamics
constants and source are excluded. Score and terminal signals are supplied
equally; collision/landing labels used only for evaluation are not added as
construction inputs. Reference dynamics and independent annotations are
accessible only to the evaluator.

\paragraph{Primary versus additional studies.}
The primary table uses four training and two development episodes per game,
each capped at 400 decisions, and evaluates control on 10 held-out seeds.
Token, interaction, and wall-time
caps are enforced independently: reaching any cap stops construction.
Control evaluation runs ten episodes per game and backbone on evaluation
seeds 1000042--1000051; the episode-level scores behind every table cell are
retained in the run records.

\paragraph{Labels and matching.}
Test annotations comprise 2,400 frames per game, with 400 double-labeled and
adjudicated. Ordinary-motion and interaction strata have fixed weights.
Object F1 uses class-consistent one-to-one IoU-$0.5$ matching; no true positives
are assigned for absent predictions. Center error is viewport-diagonal
normalized and accompanied by matched coverage. All-empty unsupported slices
are N/A. Event macro-F1 uses fixed collision, landing, score-change and
termination categories where supported, positive counts, and one-observation
temporal tolerance. Birth/death and camera slices are defined from labels,
not selected after seeing model errors. Prediction uses common held-out action
trajectories, with one inference at the anchor and no future-image updates.

\paragraph{Rollout verdicts.}
The $h_{k}$ verdict is
computed on open-loop rollouts: per class, a maximum-cardinality one-to-one
matching at IoU $\geq 0.5$ pairs predicted boxes with recorded boxes.
Matched pairs are true positives; unmatched predicted boxes are charged as
false positives and unmatched recorded boxes as false negatives, so a
copy-last-frame candidate cannot score perfectly on a moving scene and
object-count mismatches are penalized rather than excluded. A window is
correct only if every class attains a complete matching.

\paragraph{Replay distance.}
The attribution feedback distance $d(\widehat y_{t+h\mid t},y_{t+h})$ used during
construction is a bounded geometry residual over Hungarian-matched boxes:
for a matched pair with truth size $(w,h)$ and center offset
$(\Delta x,\Delta y)$ (normalized by $2w,2h$ respectively),
\begin{equation}
 \begin{aligned}
 d=\tfrac{1}{|\mathcal M|}\sum_{\mathcal M}\Bigl[&\mathcal H\!\left(\tfrac{\Delta x}{2w}\right)
 +\mathcal H\!\left(\tfrac{\Delta y}{2h}\right)
 +\tfrac14\mathcal H\!\left(\ln\tfrac{w'}{w}\right)\\
 &+\tfrac14\mathcal H\!\left(\ln\tfrac{h'}{h}\right)
 +\tfrac14\bigl(1-\operatorname{IoU}\bigr)\Bigr],
 \end{aligned}
 \label{eq:replaydistance}
\end{equation}
where $\mathcal H$ is the Huber loss ($\mathcal H(v)=\tfrac12v^2$ for $|v|\leq 1$, $|v|-\tfrac12$
otherwise), $w',h'$ are the predicted sizes, $\mathcal M$ is the matched set, and
available score and event fields are compared as auxiliary terms with the
same weighting as in the selection objective; unavailable fields are masked.
Unmatched births and deaths carry no penalty (object count is handled by the
verdict above), and a window with no comparable boxes keeps its baseline
residual rather than being rewarded with zero.

\section{Per-Game Results, Uncertainty, and Failures}
\label{app:pergame}

\subsection{Normalized Control Returns}
\label{app:raw-control}
Table~\ref{tab:control-normalized} reports the normalized counterpart of
Table~\ref{tab:control} in the main text: per-game returns are mapped to
$Z$ with the same frozen Random and Ref anchors, and rows are ordered by
backbone family. The equal-game means match Table~\ref{tab:control}.

% Normalized returns derived from control_table_data.json (raw means and
% frozen random/oracle anchors; equal-game means from the same record).
\begin{table*}[t]
\centering\fontsize{9.5}{11.5}\selectfont\setlength{\tabcolsep}{3pt}
\caption{\textbf{\method{} normalized control returns across model backbones and games.} Normalized
return (Random $=0$, Reference $=1$);
Reference denotes fixed privileged-state controllers.}
\label{tab:control-normalized}
\begin{tabular}{lrrrrrrrrr}
\toprule
\raisebox{5pt}{Backbone} & \shortstack{Chrome\\Dino} & \shortstack{Flappy\\Bird} & \shortstack{Doodle\\Jump} & \shortstack{Happy\\Hop} & \raisebox{5pt}{Asteroids} & \shortstack{Fruit\\Catcher} & \shortstack{Road\\Fighter} & \shortstack{Little\\Ski} & \raisebox{5pt}{Mean}\\
\midrule
Random & 0.000 & 0.000 & 0.000 & 0.000 & 0.000 & 0.000 & 0.000 & 0.000 & 0.000 \\
Ref & 1.000 & 1.000 & 1.000 & 1.000 & 1.000 & 1.000 & 1.000 & 1.000 & 1.000 \\
\midrule
Qwen3.8 Flash & $-0.072$ & 0.000 & 0.087 & 0.098 & $-0.066$ & 1.531 & $-0.055$ & 0.236 & 0.220\\
Kimi K3 & $-0.065$ & 0.000 & 0.174 & 0.000 & 0.317 & 1.853 & 0.001 & 1.244 & 0.441\\
GLM-5.3-Flash & $-0.075$ & 0.000 & 0.165 & 0.469 & 0.296 & 1.256 & $-0.027$ & 1.499 & 0.448\\
MiniMax-M3 & $-0.075$ & 0.050 & 1.175 & 0.716 & 0.394 & $-0.063$ & $-0.171$ & 0.652 & 0.335\\
MiniMax-M3.1-Flash & 0.684 & 0.000 & $-0.026$ & 0.098 & 0.156 & 1.106 & $-0.162$ & 1.349 & 0.401\\
GPT-6 Luna & 0.736 & 0.300 & 0.249 & 0.528 & 0.421 & 1.252 & 0.261 & 0.716 & \textbf{0.558}\\
GPT-6.1 Sol & 0.955 & 0.575 & 1.438 & 0.000 & 0.667 & 0.389 & 0.459 & 1.161 & \underline{0.706}\\
DeepSeek V4.1 Flash & 0.240 & 0.025 & 0.066 & 0.083 & 0.323 & 1.204 & 0.300 & 0.154 & 0.299\\
\bottomrule
\end{tabular}
\end{table*}

\section{Development Cost and Online Computation}
\label{app:cost}

\subsection{Direct VLM Agent Configuration}
\label{app:direct-vlm-agent}
The direct VLM baseline reuses the GameWorld agent-harness
configuration~\cite{ouyang2026gameworld} on our eight games: the
game is paused during inference, so scores reflect decision quality
rather than reaction speed, exactly as in GameWorld's paused
evaluation protocol. At every decision step the agent receives the
current screenshot, the declared game description $d_{g}$, and the same
native action list used by \method{}'s controllers, including the
\texttt{move(x,y)}/\texttt{click(x,y)} pointer verbs on mouse-driven
games. The model first reasons freely about the observation, may
request up to three zoomed crops of the screenshot for close
inspection before answering, and ends its reply with one native
action, which is executed after the game resumes; the episode then
advances by one decision step. Episodes are capped at 400 decisions,
matching the \method{} evaluation protocol, and the ten evaluation
episodes per game run on the same seeds 1000042--1000051. Failed VLM
calls are retried up to three times per step. Latency reported in
Table~\ref{tab:direct-vlm-cost} is the per-step wall-clock time of
this loop, including zoom round-trips.

\subsection{Direct VLM Agent Cost Comparison}
\label{app:direct-vlm-cost}
Table~\ref{tab:direct-vlm-cost} (main text) compares the Direct VLM agent with
\method{} under the DeepSeek V4.1 Flash backbone on the eight-game suite. The
agent selects native actions from screenshots at each decision step
and accumulates input tokens throughout ten evaluation episodes per game;
\method{} spends a one-time construction budget per game and executes
evaluation with no further model calls. The agent pays roughly three times
the total input tokens (12.2M versus 4.06M) and about 1.6 times the
total workflow time (Table~\ref{tab:direct-vlm-total-time}), while scoring below the compiled CWM on
all eight games (closest on Flappy Bird: 0.0 versus 0.1). Per-episode cost amortizes differently: a compiled model
can be re-run at near-zero marginal cost, whereas the agent pays the
perception-action loop cost again in every session.

\begin{table}[t]
\centering\fontsize{9.5}{11.5}\selectfont
\caption{\textbf{Total workflow time (seconds) over ten evaluation episodes.}
For \method{}, time includes one-time construction and evaluation.}
\label{tab:direct-vlm-total-time}
\begin{tabular}{lrr}
\toprule
Game & Direct VLM & \method{} \\
\midrule
Fruit Catcher & 4744 & 1706 \\
Chrome Dino & 7137 & 1066 \\
Doodle Jump & 3647 & 5219 \\
Happy Hop & 2753 & 3249 \\
Flappy Bird & 2520 & 1756 \\
Asteroids & 4038 & 1580 \\
Road Fighter & 2473 & 983 \\
Little Ski & 651 & 1901 \\
\midrule
Total & 27963 & 17460 \\
\bottomrule
\end{tabular}
\end{table}

\section{Implementation Map and Reproduction Checklist}
\label{app:notation}
Observer construction maps to \path{components/observer/}; causal state
processing to \path{components/observer/state_pipeline.py}; runtime validation to
\path{core/predictor_runtime.py}; candidate development to
\path{components/predictor/workbench/}; selection diagnostics to
\path{components/predictor/worldid/metrics.py}; numeric search to
\path{components/predictor/workbench/numerics.py}; and orchestration to
\path{core/co_evolution.py}.

A reproducible run bundle contains the exact input description, game and
adapter digest, model identifier, checkpoint digests, all split seeds,
annotation rules, candidate versions, prompt/tool transcript, parameter
bounds and search trace, Observer and controller hashes, failed attempts,
resource counters, and paired test outputs. Test evaluation cannot trigger
artifact revision. A test set exposed during development is retired and
replaced. All reported summary values must be regenerable from those bundles;
a hand-authored result table is not a substitute for the record.

\section{Open-Loop Prediction Across Horizons}
\label{app:curves}
\begin{figure*}[t]
\centering
\includegraphics[width=\textwidth]{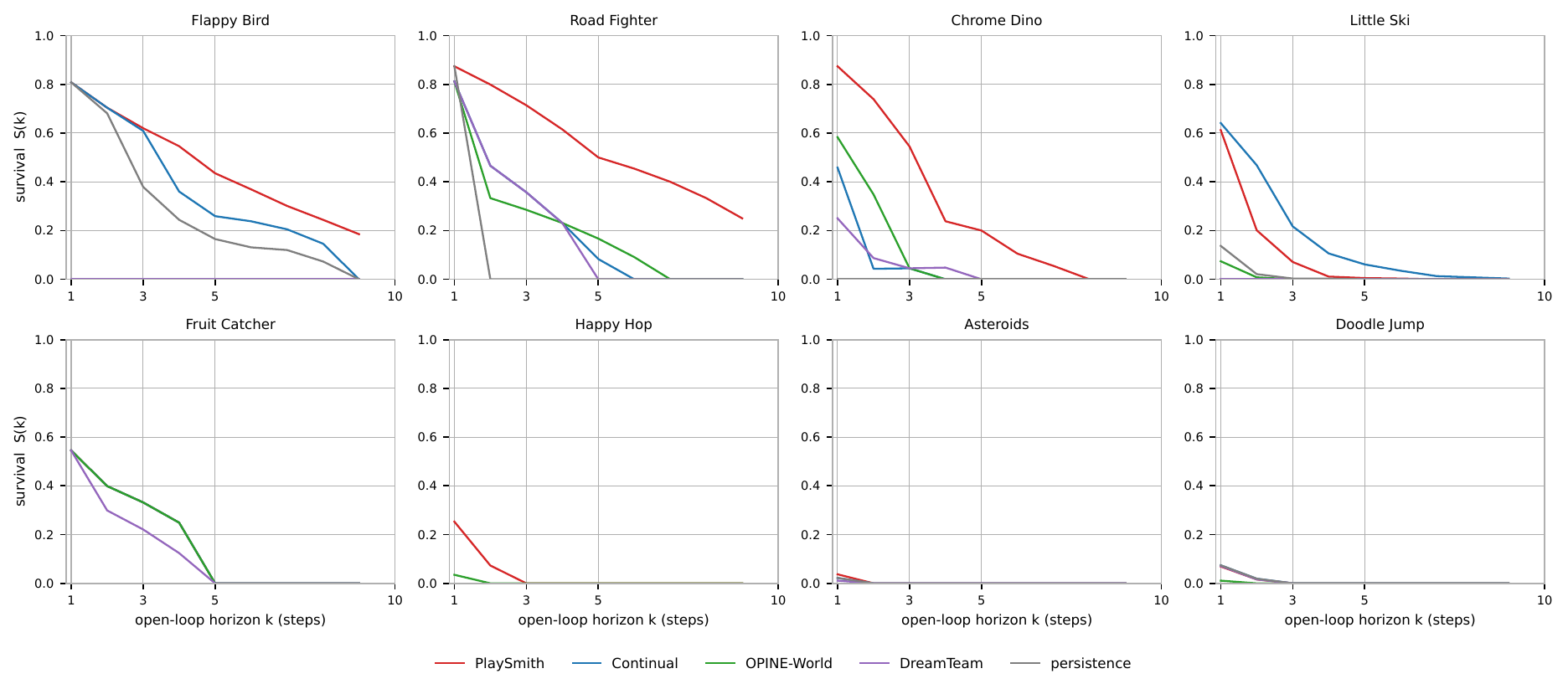}
\caption{Open-loop survival $S(k)$ across horizons on all eight games. \method{} (red) versus the transplanted
construction procedures and the persistence floor.}
\label{fig:curves}
\end{figure*}

\begin{figure*}[t]
\centering
\includegraphics[width=\textwidth]{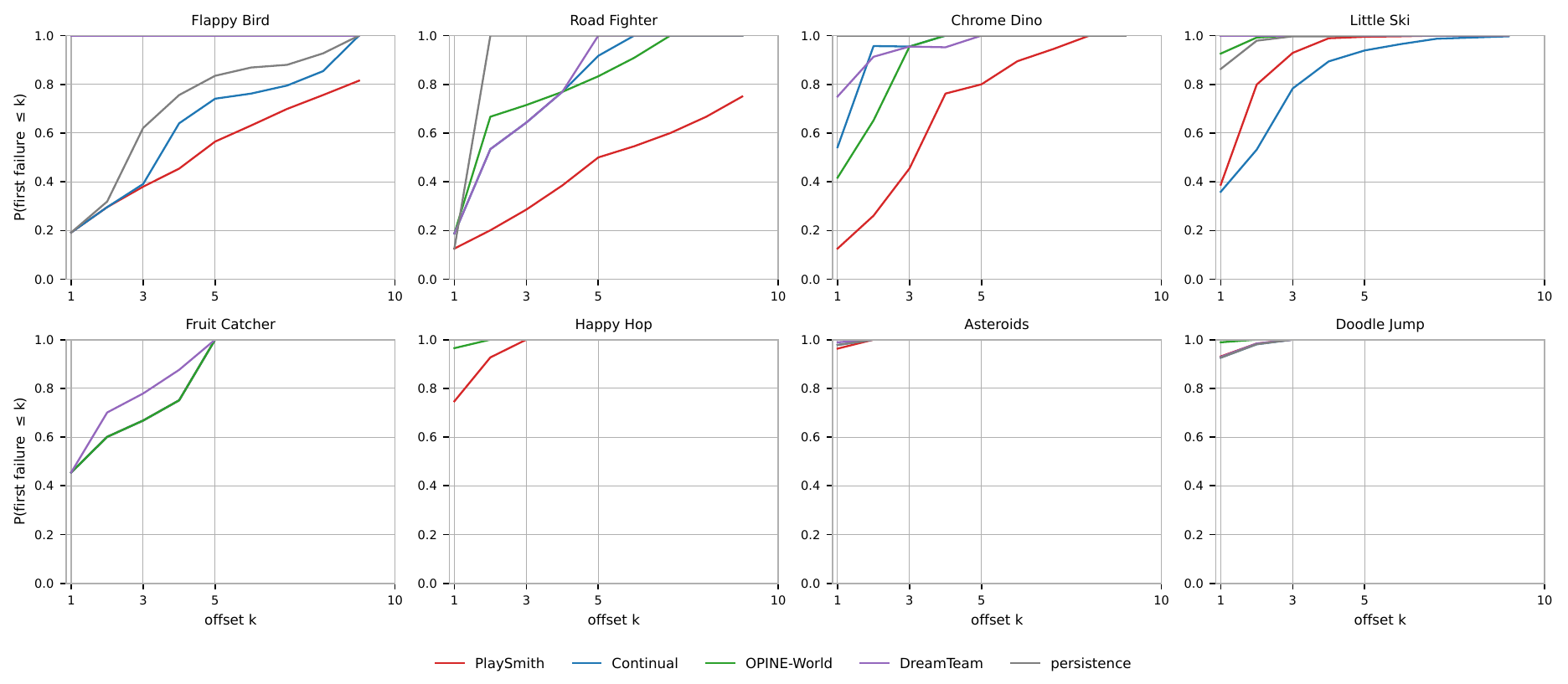}
\caption{First-failure CDF $1-S(k)$ on all eight games. Lower-right is
better (errors appear later). The last four games bound the regime: all
constructors fail by $k{=}3$--$5$ there.}
\label{fig:curves-cdf}
\end{figure*}

All frozen models from the construction-method comparison
(Table~\ref{tab:construction-methods}) are replayed on the common test trajectories
with one inference at the anchor and no future-image updates, using the
same replay instrument as construction-time development. For each anchor,
$S(k)$ is the fraction of anchors whose entire open-loop rollout remains
correct through offset $k$ (strict joint geometry/count criterion;
Fig.~\ref{fig:curves}). Fig.~\ref{fig:curves-cdf} shows the
complementary view: the cumulative probability that an anchor's first
mismatch occurs at or before offset $k$, i.e.\ $1-S(k)$. Curves in the
lower-right of Fig.~\ref{fig:curves-cdf} fail later.
On the four games where any constructor passes the first step, the
\method{} Predictor's advantage over the strongest alternative grows with
horizon (e.g.\ Flappy Bird: equal $S(1)$, $S(5)=0.44$ versus $0.26$); the
remaining games bound the regime: every constructor collapses to the floor
before $k{=}5$ there, so no horizon effect is measurable.

\section{Trajectory Evidence and Predictive-State Hypotheses}
Associated object histories provide displacement rates in explicit coordinates
and sampling units. Track identities take precedence over geometric matching;
ambiguous pairs, missing detections and episode boundaries terminate a track.
Action-stratified summaries retain their sample counts and temporal provenance.
An empty sample denotes unavailable evidence rather than zero change.
Local trajectory estimates supply additional evidence for candidate state fields.
For example, similar Doodle platform positions can have opposite historical
displacements; an inferred scene velocity distinguishes their future evolution.
History-dependent changes can also motivate a phase or timer. The VLM proposes
these fields and their update rules, and the loop evaluates the resulting rollouts.
Changing constants in a fixed representation does not alter its available memory;
revising state inference and transition code can preserve pending actions or events.

\section{Predictor Construction Loop and Auxiliary Procedures}
\label{app:predictorprocedures}
\paragraph{Tool execution and candidate records.}
The construction loop offers training measurements, image and source inspection,
and optional bounded numerical search. Its VLM policy $\mathcal V$ selects
operation $u_{k}$ from game description $d_{g}$ and development transcript $\eta_{k}$:
\begin{equation}
 \begin{split}
 u_{k}&\sim\mathcal V(\cdot\mid d_{g},\eta_{k}),\\
 v_{k}&=\operatorname{ExecuteTool}(u_{k};\mathcal D_{\mathrm{tr}},\mathcal D_{\mathrm{dev}}).
 \end{split}
 \label{eq:toolloop}
\end{equation}
Tool results enter $\eta_{k+1}$; validated submissions become versioned
candidates. New environment interaction follows the run's information contract.

\paragraph{Replay diagnostics and auxiliary numerical fitting.}
The following objectives specify diagnostic summaries and the bounded
constant-fitting facility used alongside Predictor program revision.
For notation, the replay evidence is a diagnostic vector
\begin{equation}
 \boldsymbol\ell=(\ell_{\mathrm{geom}},\ell_{\mathrm{count}},\ell_{\mathrm{score}},
 \ell_{\mathrm{event}},\ell_{\mathrm{run}}).
 \label{eq:diagnostics}
\end{equation}
The harness versions its candidate-selection score separately from independent
evaluation metrics.
Over valid anchor/horizon windows $\mathcal W_{\mathcal D}$, summarize replay
evidence by the componentwise masked mean
\begin{equation}
 \boldsymbol L_{\mathrm{rep}}(p,\theta;\mathcal D)=
 \operatorname{MaskedMean}_{(\tau,t,h)\in\mathcal W_{\mathcal D}}
 \boldsymbol\ell(\widehat y_{t+h\mid t},y_{t+h}).
 \label{eq:replayrisk}
\end{equation}
Each prediction is generated by the candidate Predictor with program
$p$ and constants $\theta$ using the Observer history through $t$ and
the recorded actions and intervals for the next $h$ steps.
The reference $y_{t+h}$ uses the recorded prefix through $t+h$.
Masks reflect unavailable reference fields, not whether the candidate predicts
them successfully. Exceptions remain failures. This diagnostic functional is
not substituted for the implementation's actual selection rule.

For a fixed program, optional numerical search examines an executed set of
parameter proposals $\Theta_{B}(p)$ using training replay:
\begin{equation}
 \theta_{p}^{\mathrm{fit}}\in\argmin_{\theta\in\Theta_{B}(p)}
 \mathcal L_{\mathrm{fit}}(p,\theta;\mathcal D_{\mathrm{tr}}).
 \label{eq:fitting}
\end{equation}
The search records its evaluated proposals. Verbatim submissions retain their
authored constants.

\paragraph{Additional collection rounds.}
The optional outer collection schedule updates the CWM's Observer and
Predictor between rounds. The loop operates within Predictor construction in each
round; the collection schedule is separate from that inner loop.
For $R$ construction rounds, initialize $\mathcal D^{(-1)}=\varnothing$ and
a declared cold-start policy $\pi^{(-1)}$. $B$ specifies round-level limits and
any run-wide cap; reported resources are cumulative. Accumulate experience by
\begin{equation}
 \mathcal D^{(r)}=\mathcal D^{(r-1)}\cup
 \operatorname{Collect}(g,\pi^{(r-1)}),\quad r=0,\ldots,R-1.
 \label{eq:rounds}
\end{equation}
The accumulated experience supports rebuilding the Observer and Predictor.

\begin{algorithm}[t]
\caption{Visual-to-executable construction scaffold}
\label{alg:collection}
\begin{algorithmic}[1]
\Require Game interface, rules $d_{g}$, budget $B$, rounds $R$
\State $\mathcal D\gets\varnothing$; initialize $\pi^{(-1)}$
\For{$r=0,\ldots,R-1$}
\State Collect training episodes with $\pi^{(r-1)}$; append to $\mathcal D$
\State Construct/update Observer and causal histories
\State Partition whole episodes into training/dev evidence
\State Develop model candidates by bounded tool execution
\If{no executable candidate has finite development score}
\State Record construction failure and stop
\EndIf
\State Select executable model using development evidence
\State Generate/update controller $\pi^{(r)}$; record hashes
\EndFor
\State Freeze artifacts; evaluate on held-out episodes
\end{algorithmic}
\end{algorithm}
% Protocol audit: trials used for revision are train/dev, not final test.
% The final freeze/evaluate step specifies the paper protocol; historical exploratory runs require separate auditing.

\section{Detailed Comparison Setups}
\label{app:comparison}
Table~\ref{tab:detailed-comparison} expands the main-text comparison
(Table~\ref{tab:related-comparison}) with observation interfaces, learned
artifacts, and execution settings of related systems.
The four marks in the main table describe the reported model-induction
interface, rather than every modality an environment can display.
Mind-Studio's synthesis uses OCAtari state-action-next-state trajectories,
with screenshot-derived skill information~\cite{dong2026mindstudio}; it
therefore does not meet the raw-pixel criterion. Tycho's actionable
history is structured ARC-AGI-3 interaction evidence~\cite{lehmann2026tycho},
and Twin's replay validation uses structured grids and action-indexed
counterexamples~\cite{skoutnev2026twin}. The latter is automatic
transition-error detection, but not the action-versus-autonomous
continuous-motion attribution used for the final column. OPINE-World
also repairs action-conditioned grid transitions~\cite{courtis2026opineworld};
its grid setting does not meet that motion criterion.
\begin{table*}[t]
\centering
% Detailed system setups extending the compact feature matrix.
% Source audit: table1_feature_audit.md; entries pending final verification.
\begingroup
\fontsize{9.5}{11.5}\selectfont\setlength{\tabcolsep}{3pt}\renewcommand{\arraystretch}{1.12}
\begin{tabularx}{\textwidth}{@{}l>{\raggedright\arraybackslash}p{.14\textwidth}>{\raggedright\arraybackslash}X>{\raggedright\arraybackslash}X@{}}
\toprule
Work & Observation interface & Learned artifacts & Execution setting\\
\midrule
WorldCoder~\cite{tang2024worldcoder} & Text state & transition program & grid replay\\
GIF-MCTS~\cite{dainese2024gifmcts} & Control states & planner + dynamics & grid control\\
PoE-World~\cite{piriyakulkij2025poeworld} & RAM features & transition program & Atari replay\\
Mind-Studio~\cite{dong2026mindstudio} & OCAtari extraction & skills + scene maps & Atari control\\
OPINE-World~\cite{courtis2026opineworld} & color-grid parser & refined grid program & ARC grids\\
Tycho~\cite{lehmann2026tycho} & grid + optional renders & world program & ARC grids\\
Twin~\cite{skoutnev2026twin} & text grids & agent + world models & ARC exploration\\
FAE~\cite{goel2026fae} & video sprites & automata program & sprite replay\\
Code-as-World~\cite{wang2026codeasworld} & video + physics & editable scene code & quantitative QA\\
\midrule
\textbf{\method} & learned detector + association & Observer + \texttt{Infer}/\texttt{Transition}/\texttt{Render} + controller & browser control\\
\bottomrule
\end{tabularx}\endgroup

\caption{Detailed system setups and capabilities of related systems.
These descriptions do not establish a performance ranking.}
\label{tab:detailed-comparison}
\end{table*}

\section{Adapted Predictor Baselines}
\label{app:adapted-baselines}
Table~\ref{tab:baseline-adaptations} records how the published ideas were
mapped to the common Predictor interface. These are new implementations
in our harness; the cited papers do not report the numbers in
Table~\ref{tab:construction-methods}. All methods receive the same
frozen Observer output, native action vocabulary, and train/development
split. A candidate must implement the same state inference, transition,
and rendering interface, and selection uses development evidence;
the held-out test split is reserved for final scoring. The construction
budget is 50k tokens per game under DeepSeek V4.1 Flash.
\begin{table*}[t]
\centering\small
\begin{tabularx}{\textwidth}{@{}l>{\raggedright\arraybackslash}p{.29\textwidth}>{\raggedright\arraybackslash}X@{}}
\toprule
Adapted method & Retained construction mechanism & Interface change and stopping rule\\
\midrule
OPINE-World~\cite{courtis2026opineworld} & Counterexample-guided whole-module revision, evidence ledger, critic, optional experiments & Replace ARC grid extraction with the frozen visual Observer; emit a Predictor module rather than a grid-world program. At most five revision rounds; failed initial validation permits two repair attempts.\\
DreamTeam~\cite{sarafian2026dreamteam} & Simulator, Prober, Strategist, Critic, and Team Leader roles; routed errors and regression replay & Observer role is the shared frozen Observer; one backbone performs all retained roles. Hidden-state replay catch-up and reset reflection are omitted. At most eight arbitration rounds; failed regressions roll back.\\
Continual~\cite{karten2026continual} & Monotone failure log, Refiner, function-level incremental patches, fallback & Replace online skill-library refinement with offline Predictor refinement from complete training episodes, starting from persistence. No online exploration; at most six refinements, with validation and development-score rollback.\\
OpenCode~\cite{opencode2025} & Terminal build/plan coding workflow & The build/plan agent pair receives the shared evidence and tools, and writes only the Predictor module under the same 50k-token budget; its frozen output is scored on the common test split.\\
\bottomrule
\end{tabularx}
\caption{Adaptation of published construction ideas to the shared Predictor task. The original systems' published scores are not transferred.}
\label{tab:baseline-adaptations}
\end{table*}

\end{document}